\documentclass{article}

\usepackage{arxiv}

\usepackage[utf8]{inputenc} 
\usepackage[T1]{fontenc}    
\usepackage{hyperref}       
\usepackage{url}            
\usepackage{booktabs}       
\usepackage{amsfonts}       
\usepackage{nicefrac}       
\usepackage{microtype}      
\usepackage{lipsum}
\usepackage{graphicx}
\usepackage{float}
\usepackage{subfig}

\graphicspath{ {./images/} }

\title{Semi-Supervised Generative Adversarial Network for Stress Detection Using Partially Labeled Physiological Data}

\author{
  \textbf{Nibraas Khan} \\
  Department of Computer Science\\
  Vanderbilt University\\
  Nashville, TN 37235 \\
  \texttt{nibraas.a.khan@vanderbilt.edu} \\
  \and
  \textbf{Nilanjan Sarkar} \\
  Department of Mechanical Engineering\\
  Vanderbilt University\\
  Nashville, TN 37235 \\
  \texttt{nilanjan.sarkar@vanderbilt.edu} \\
}

\begin{document}
\maketitle
\begin{abstract}
Physiological measurements involves observing variables that attribute to the normative functioning of human systems and subsystems directly or indirectly. The measurements can be used to detect affective states of a person with aims such as improving human-computer interactions. There are several methods of collecting physiological data, but wearable sensors are a common, non-invasive tool for accurate readings. However, valuable information is hard to extract from the raw physiological data, especially for affective state detection. Machine Learning techniques are used to detect the affective state of a person through labeled physiological data. A clear problem with using labeled data is creating accurate labels. An expert is needed to analyze a form of recording of participants and mark sections with different states such as stress and calm. While expensive, this method delivers a complete dataset with labeled data that can be used in any number of supervised algorithms. An interesting question arises from the expensive labeling: how can we reduce the cost while maintaining high accuracy? Semi-Supervised learning (SSL) is a potential solution to this problem. These algorithms allow for machine learning models to be trained with only a small subset of labeled data (unlike unsupervised which use no labels). They provide a way of avoiding expensive labeling. This paper compares a fully supervised algorithm to a SSL on the public WESAD (Wearable Stress and Affect Detection) Dataset for stress detection. This paper shows that Semi-Supervised algorithms are a viable method for inexpensive affective state detection systems with accurate results.

\end{abstract}


\section{Introduction}
A humans actions, behaviors, or thoughts can be heavily influenced by emotion through changes in psychology or physiology. In other words, emotions serve as the primary motivational system for human beings \cite{izard2013human}. These emotions can be used to improve Human-Computer Interactions by adapting based on the emotions exhibited by the human. To improve interactions thorough emotion, an assumption needs to be made: There are clear and objective physiological measurements in responses to the autonomic nervous system activity which can be used for understanding emotion \cite{kreibig2010autonomic}. 

This paper defines physiological measurements as the observation of variables that relate to the normative functioning of the biological systems and subsystems for a human. Examples of such measurements are heart rate and blood pressure \cite{Myroniv2017}. Physiological data can be non-invasively measured through wearable devices which are capable of collecting signals such as such as Blood Volume Pulse signal (BVP), Heart Rate Variability (HRV), Interbeat Interval (IBI), Electrodermal Activity (EDA), Accelerometer (ACC), Thermometer (TEMP), and Heart Rate (HR) \cite{saganowski2020consumer}. These signals allow us to detect emotion \cite{9156096, s19112509, setiawan2018framework, zhang2021corrnet, ALI202123} and stress \cite{sano2013stress, giannakakis2019review, montesinos2019multi, jebelli2018supervised}. However, going from raw physiological data to understanding emotion or stress requires techniques such as Machine Learning (ML). ML techniques such as K-Nearest Neighbour, Linear Discriminant Analysis, Random Forest, Decision Tree, AdaBoost and Kernel Support Vector Machine, and Deep Learning are used \cite{9183244, khowaja2021toward, oskooei2021destress} in current research.  

Using these ML models can be expensive as the supervised learning require a large amount of high quality labeled data. Creating this high quality data could involve experts to analyze physiological measurements recorded with video or audio and label sections with affective states. For example, a one hour recording of a subject performing tasks can be broken down into ten second labeled segments. An expert could analyze each ten second segment and label for supervised learning models. It should be noted that there are other techniques of labeling physiological data such marking entire large segments as a state. For example, a participant might be asked to perform mental math for 10 minutes in the one hour recording, and that segment can be labeled as stress. This method of labeling was used for the WESAD public dataset \cite{10.1145/3242969.3242985}. Labeled datasets have been using successfully for supervised stress and affect detection \cite{Zontone2020, Li2020, Uma.S.ParlapalliPrudhvi2020}. On the other side, there have been successful for unsupervised stress and affect detection \cite{oskooei2019destress, 5b46f99bbc2542468311a59d4e8f3562, 10.1145/3395035.3425191}.

Supervised learning allows for highly accurate emotion classification or stress detection using high quality labeled data. Unsupervised Learning works primarily by clustering with unlabeled data. The expensive labeling process is one of the main limitations of fully-supervised ML algorithms. The example of the labeling process above highlights how time-consuming and expensive it is. Having a fully labeled dataset is expensive; however, labeling a small portion of the dataset is inexpensive and can still provide value. If the small, labeled portion of the dataset were to be used for supervised learning, the algorithms struggle to achieve high performance. In real world problems, unlabeled data is abundant and easily available while labeled data is expensive to generate. Therefore, there is active research into how to leverage a large amounts of unlabeled data with a small number of labeled samples known as Semi-Supervised Learning (SSL) \cite{Yang2021}. Semi-Supervised learning is being used in a wide range of applications due to its benefits of cheaper data \cite{berthelot2019mixmatch, olsson2021classmix, protopapadakis2021stacked}. 

Naturally, the question arises on how we can apply SSL for stress or emotion detection with physiological data from wearable sensors. This area of research is active with many SSL algorithms. Some SSL techniques involve stacked Denoising Autoencoders (SDAE) \cite{10.5555/1756006.1953039} and Deep Belief Networks (DBN) \cite{hinton2006fast} for extracting features from physiological data \cite{xu2016affective}. Other work has focused on Co-Training \cite{4284821, schels2014using, zhang2014cooperative} and Self-Training \cite{zhang2011unsupervised, zhang2014cooperative, 10.1145/973264.973268} and an enhancement on the techniques \cite{7472666}. The mentioned research was tested on other public datasets including the RECOLA multimodal database \cite{6553805}. While the research presented above is valuable and interesting, focus can be spent of studying generative models for physiological data. There is research using generative models \cite{xu2016affective}, but there is more to be explored with techniques such as Semi-Supervised Generative Adversarial Networks (SGAN).

This paper aims to expand the understanding of SSL techniques on physiological data by analyzing and comparing a fully-supervised learning algorithm using Bi-Directional Long-Short Term Memory (Bi-LSTM) and a semi-supervised learning algorithm using Semi-Supervised Generative Adversarial Networks (SGAN). The public WESAD dataset containing labeled physiological data collected using a wrist sensor, Empatica E4, is used for testing and comparing. This paper will also slightly highlight using Long-Short Term Memory and Convolutional Neural Networks as the inner models and feature engineering.

A successful SSL algorithm for stress detecting using physiological data will improve Human-Computer Interactions without the expensive cost of labeling. It will provide benefits in all areas that work with physiological data and ML. It is key to leverage partially labeled data with abundantly available unlabeled data to create powerful inexpensive emotion detection systems to improve Human-Computer Interactions and other related fields. 

The results of this study show that SGANs are a potential solution to this problem. When compared to a fully-supervised system, it was not able to perform as well. However, even with limited data, it was able to solve the detection problem. SSL algorithms provide a solution for expensive labeling with physiological data.

\section{Methods}

\begin{figure}[ht]
\includegraphics[scale=0.2]{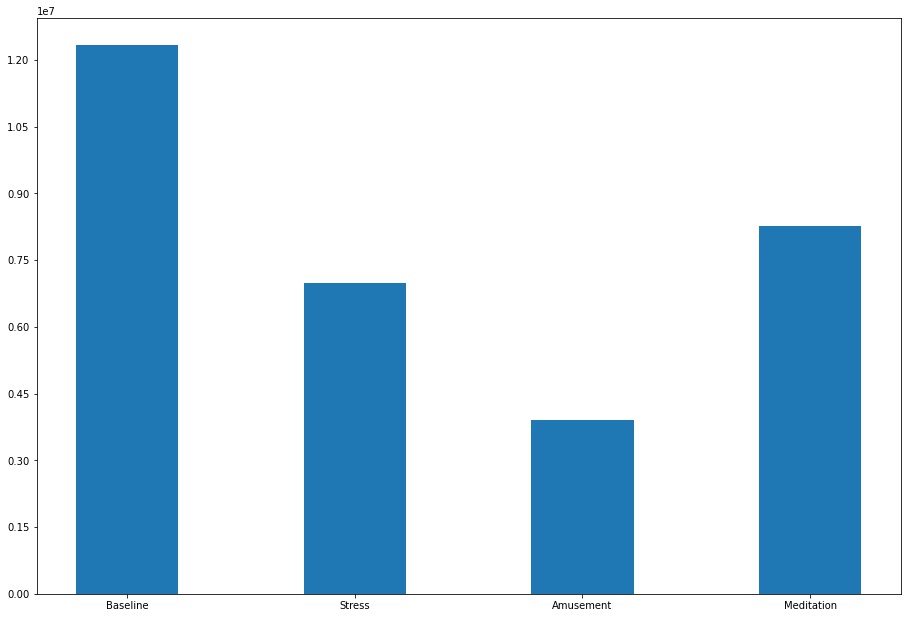}
\centering
\caption{Count of labels (Baseline, Stress, Amusement, and Meditation) in the WESAD dataset.}
\label{fig:labels}
\end{figure}

\begin{figure}[ht]
\includegraphics[scale=0.2]{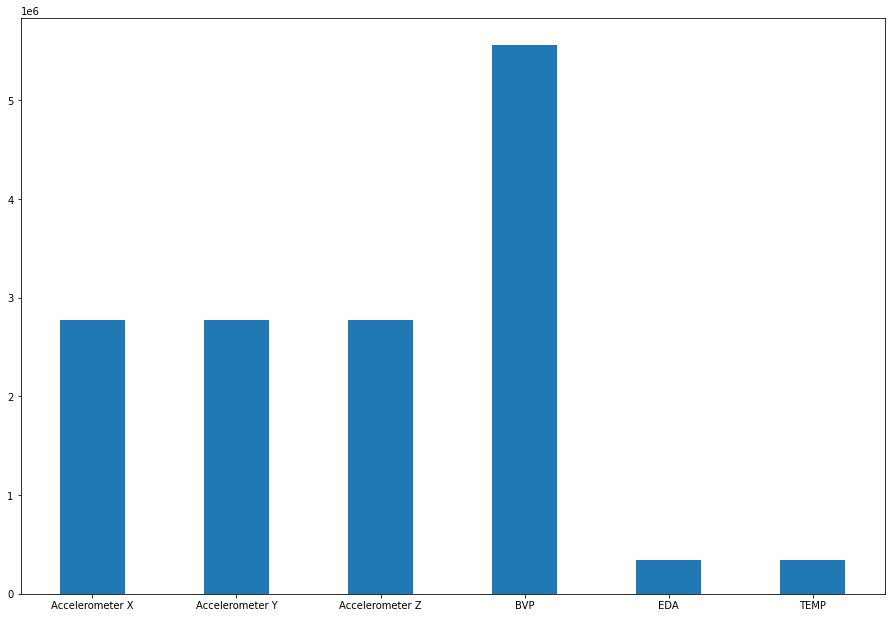}
\centering
\caption{Amounts of collected data (ACC, BVP, EDA, and TEMP) in the WESAD dataset.}
\label{fig:number_of_data}
\end{figure}

\begin{figure}
    \centering
    \subfloat[Accelerometer X-axis]{\includegraphics[width=0.24\textwidth]{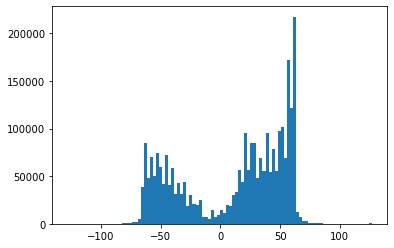}} 
    \subfloat[Accelerometer Y-axis]{\includegraphics[width=0.24\textwidth]{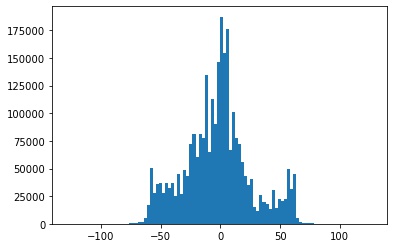}} 
    \subfloat[Accelerometer Z-axis]{\includegraphics[width=0.24\textwidth]{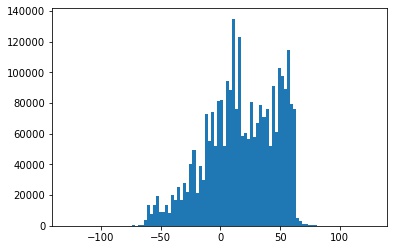}}\\
    \subfloat[BVP]{\includegraphics[width=0.24\textwidth]{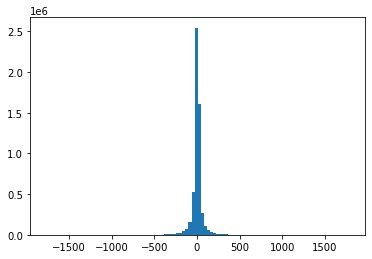}}
    \subfloat[EDA]{\includegraphics[width=0.24\textwidth]{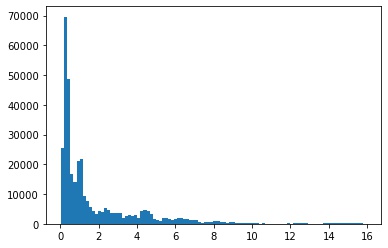}}
    \subfloat[TEMP]{\includegraphics[width=0.24\textwidth]{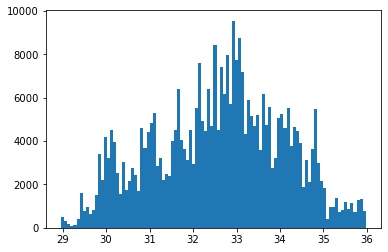}}
    \caption{Histogram of the different data types collected by the E4 sensor.}
    \label{fig:histo}
\end{figure}

\subsection{Data}
The WESAD dataset is a public dataset hosted by the University of California Irvine for wearable stress and affect detection. The dataset contains multimodal physiological and motion data from 15 participants using both a wrist and chest device (Empatica E4 and RespiBAN, respectfully). The data generated by the RespiBAN are: Electrocardiogram, Electrodermal Activity, Electromyogram, Respiration, Body Temperature, and three-axis Acceleration sampled at 700 Hz. The data generated from the E4 sensors are: Blood volume pulse (BVP, 64 Hz), electrodermal activity (EDA, 4 Hz), body temperature (4 Hz), and three-axis acceleration (32 Hz). Each participant also has self-reported stress scores based on questionnaires \cite{10.1145/3242969.3242985}.

The authors of the dataset provide four ground truth labels: Baseline, Stress, Amusement, and Meditation. These are considered ground truth based on what the participants were asked to do during the time of when the labels were collected. For example, the participants were asked to perform a stressful task such as performing mental math calculations, and that time they spent working on it was recorded as stress. The number of labels of the WESAD data are shown in Figure \ref{fig:labels}. The figure highlights the number of labeled for the entire WESAD dataset as set by the authors. A large portion of the data is labeled as Baseline, which is just used to normalize the participants data. 

For the purposes of this paper, only the wrist sensor, Empatica E4, was considered and the task was a binary classification problem of whether the participant was stressed or not stressed. The baseline label was used to normalize the subject data, and the binary classification was formatted as meditation and amusement is not stressed and stress is marked as stress. 

The multi-model dataset contains motion data (ACC) and physiological data (BVP, EDA, and TEMP). Figure \ref{fig:number_of_data} shows the number of samples of each of the data points collected by the E4. The spread of each of the motion and physiological data is shown as histograms in Figure \ref{fig:histo}. The WESAD dataset serves as a perfect dataset to use for this paper as it contains multi-model information for stress classification and contains data from a popular wearable sensor. The sections below talk about the data along with the models used. 

\subsection{Problem}
By analyzing Figure \ref{fig:number_of_data}, it is clear that the E4 sensor collects large amounts of data. Creating a supervised model of affect detection using this data requires a lot of expert labeling. The labeling can be done the same way the WESAD authors did or the data can be broken down into smaller segments and labeled manually. The second method provides higher quality labels, but it is expensive. 

An SSL algorithm can be used to alleviate the expenses of labeling while leverage small amounts of labeled data. This method differs from unsupervised learning as some labeled data will be used to guide the model. A successful SSL algorithm would allow for improvements in Human-Computer Interactions without the need of expensive, time-consuming labeling.

This paper aims to analyze the effectiveness of using SSL algorithms for stress detection using physiological data collected from wearable Empatica E4. A Semi-Supervised Generative Adversarial Network is compared with a fully supervised Bi-Directional LSTM Network. Creating both a supervised and an SSL algorithm, allows for comparison and analysis. A successful SSL algorithm will be able to perform close to the supervised algorithm and should continue to increase in performance as the ratio of labels increases.

\subsection{Data Preparation}
The focus of this paper was comparing the two model. Therefore, the feature engineering was made to be as clear and basic as possible. There were two methods considered for preparing the data:

\begin{itemize}
  \item No Preparation - Use the raw data and labels provided by the E4 sensor
  \item Basic Preparation - Use the mean, max, min, range, and standard deviation with a rolling window
\end{itemize}

The idea of using an end-to-end ML model is a popular area of research with success \cite{Dziezyc2020, 10.1145/3341163.3347741}. The first models were tested using the No Preparation method, and the results were comparable to Basic Preparation method. For the physiological data, both No Preparation and Basic Preparation methods are not significantly different for the purposes of this work.

In practice, the deciding factor came to the running time of the algorithms. Both the techniques were used while building the supervised model, and it was clear that the No Preparation method took significantly longer to train compared to the Basic Preparation. The time difference is due to the number of data points used in training. In the Basic Preparation method, a rolling window of size 42000 (60 seconds of recorded data according to the authors of WESAD) is used to calculate mean, max, min, range, and standard deviation of all the data types collected by the sensor. 

Specifically, the steps for Basic Preparation are as follows for each subject:

\begin{itemize}
  \item Normalize each participants data using the baseline label (subtract all the data by the mean of the baseline)
  \item Grab a window of size 42000 with a step size of 175 (60 seconds of data)
  \item Impute missing data with the mean of the window for each type of collected data 
  \item ACC - Split the data into x, y, and z axis and calculate the min, max, mean, range, and standard deviation of the current window for each axis
  \item BVP, EDA, and TEMP - Calculate the min, max, mean, range, and standard deviation of the current window for each type of collected data
\end{itemize}

The results of testing the two techniques led to using the Basic Preparation method for both the supervised and SSL technique.

\subsection{Proposed Models}

One of the most important factors when developing the two algorithms is what kind of Neural Network to use. There were three models that were considered, and this section will highlight them and explain why the Bi-directional LSTM Model won.

\subsubsection{Long-Short Term Memory}
The first model considered was a Long-Short Term Memory Network. Since the physiological data is time-series or sequential, this model is of interest. Recurrent Neural Networks and LSTMs are effective and scalable models for several learning problems related to sequential data \cite{Hochreiter95longshort-term, 7508408, graves2005framewise, gers1999learning, gers2000recurrent}. 

\begin{figure}[ht]
\includegraphics[scale=0.4]{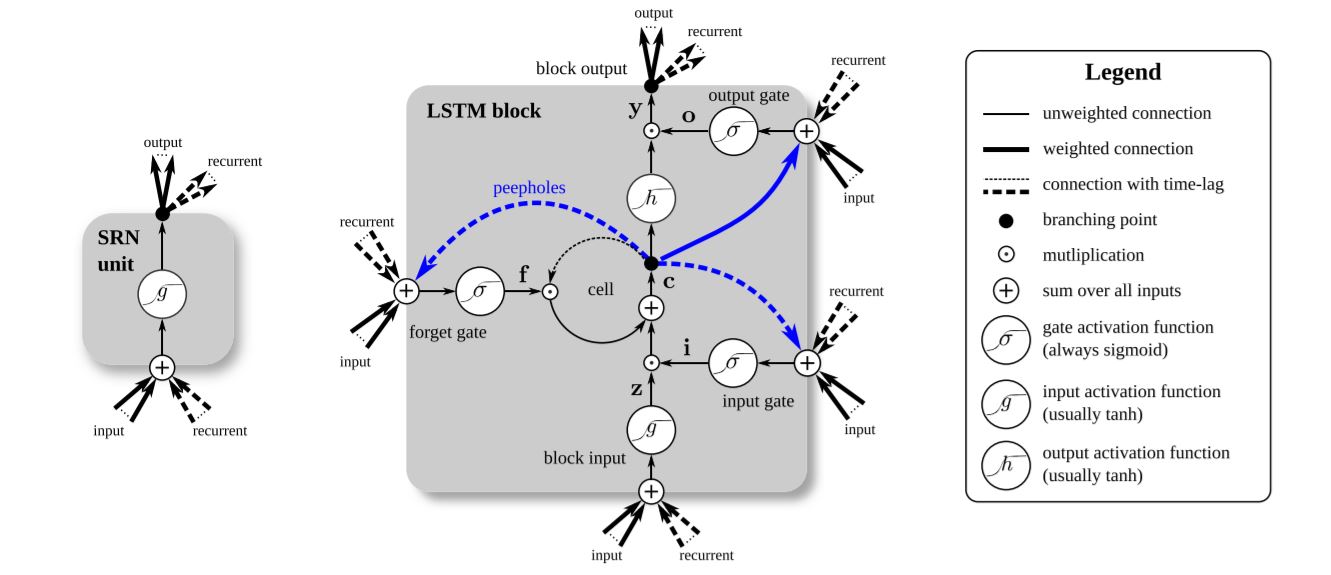}
\centering
\caption{The left shows the schema of a simple RNN and the right shows the schema of a LSTM network. This image is used from the works of Greff et al. \cite{7508408}.}
\label{fig:lstm_model}
\end{figure}

The common vanilla LSTM modules used in research contain three gates (input, forget, and output), block input, a single cell (the constant error carousel), an output activation function, and peephole connections. The recurrence aspect of the model is created by connecting the output block back to the block input and the gates.

While the details of the inner workings of the LSTM model are out of scope for this paper, a short visualization is shown in Figure \ref{fig:lstm_model}.

\subsubsection{Convolutional Neural Networks}

Research in physiology has found that Convolutional Neural Networks (CNNs) might perform better on physiological data collected from wearable senors. Dzie{\.{z}}yc et al. tested Fully Convolutional Network \cite{wang2018multilevel}, Time Convolutional Neural Network \cite{zhang2016understanding}, Multichannel Deep Convolutional Neural Network \cite{zhang2016understanding}, and Convolutional Neural Network with Long-Short Term Memory \cite{kanjo2019deep} along with other techniques such as LSTM and found that CNN based models might be better suited than LSTM based models for physiological data \cite{Dziezyc2020}.

\begin{figure}[ht]
\includegraphics[scale=0.6]{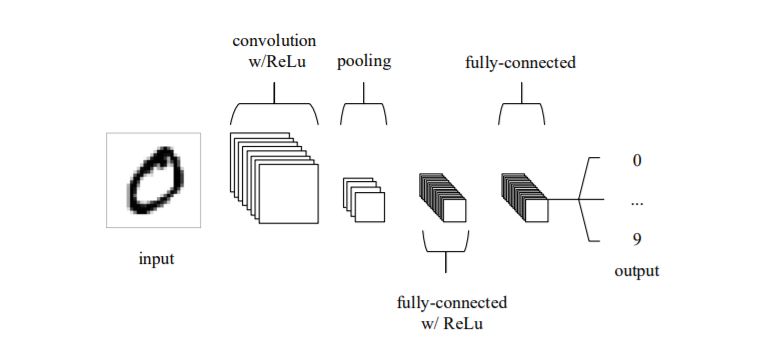}
\centering
\caption{Simple example of a CNN model used to solve the MNIST dataset. This image is used from the works of O'Shea et al. \cite{oshea2015introduction}.}
\label{fig:cnn_model}
\end{figure}

The considered CNN model was a simple model that contained three types of layers: convolutional layers, pooling layers and fully-connected layers. An example of a CNN model used for a sample problem (handwritten digit prediction on the MNIST dataset) is shown in Figure \ref{fig:cnn_model}.

\subsubsection{Bi-Directional Long-Short Term Memory}
The purpose of the LSTM model is preserve information from inputs that has already passed by using the hidden state. It works well with sequential data since it stores sequential information in its hidden states. The standard LSTM model preserves information of the past since its inputs are only things seen in the past.

The Bi-Directional LSTM will allow the model to be trained with inputs in two ways: past to the future and future to the past. Using this model allows for the preservation of both information from the future and the past. 

In short, the standard LSTM model learns context from the data in one direction (forward), while the bidirectional LSTM model learns context by passing the data both forward and backwards.

While some research suggests the CNN work better than LSTM models, most of the research on classification of affect states using physiological data uses LSTM models. Since the focus of the paper is to compare standard supervised to SSL, the Bidirectional LSTM model was chosen. Bidirectional was chosen over the standard LSTM model to allow the model to learn as much context from the data as possible. 

\subsection{Supervised Learning}

\begin{figure}[ht]
\includegraphics[scale=0.4]{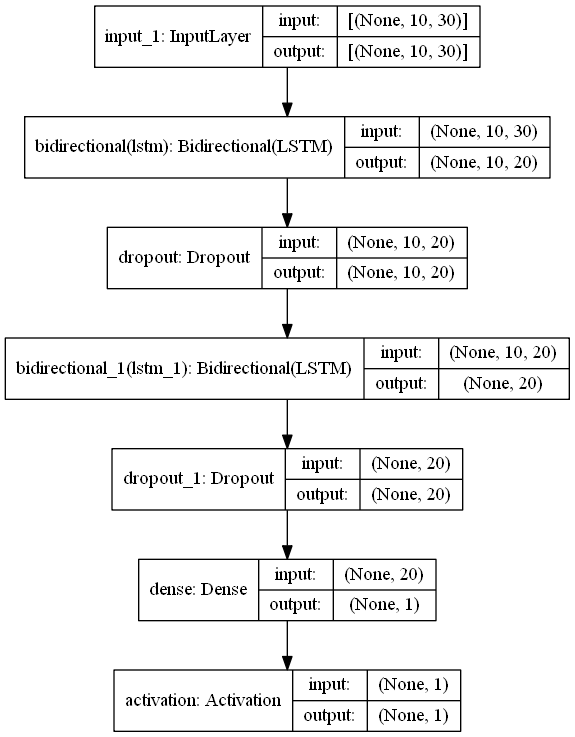}
\centering
\caption{Schema of the Supervised Learning algorithm generated by Keras.}
\label{fig:kerassup}
\end{figure}

Figure \ref{fig:kerassup} shows the schema of the Bi-Directional LSTM model used in this paper. The model had two hidden Bi-Directional LSTM layers of size 10 connected to dropouts with a probability of 0.2. The output is a single node with a sigmoid activation function. The input of the model was 30 features with 10 steps in the past. The model looked at 10 data points in the past to predict the stress value. 

The schema presented above was the final model chosen, but there were several iteration before. There were three main model that were considered: a smaller network with 5 LSTM nodes, the model presented above, and a model with 50 LSTM nodes. It was clear that the smaller model was not able to learn with 5 nodes, and the model with 50 nodes produces dramatically different results every time it is run since the search space is so large. So the middle ground model above was chosen.

\subsection{Semi-Supervised Generative Adversarial Networks}

The SSL algorithm used in this paper is a generative adversarial network. To keep comparisons as fair as possible, the discriminator and the classifier of the network were built using the same schema as the fully supervised learning algorithm.

\begin{figure}[H]
    \centering
    \subfloat[Shared Discriminator and Classifier Model]{\includegraphics[width=0.24\textwidth]{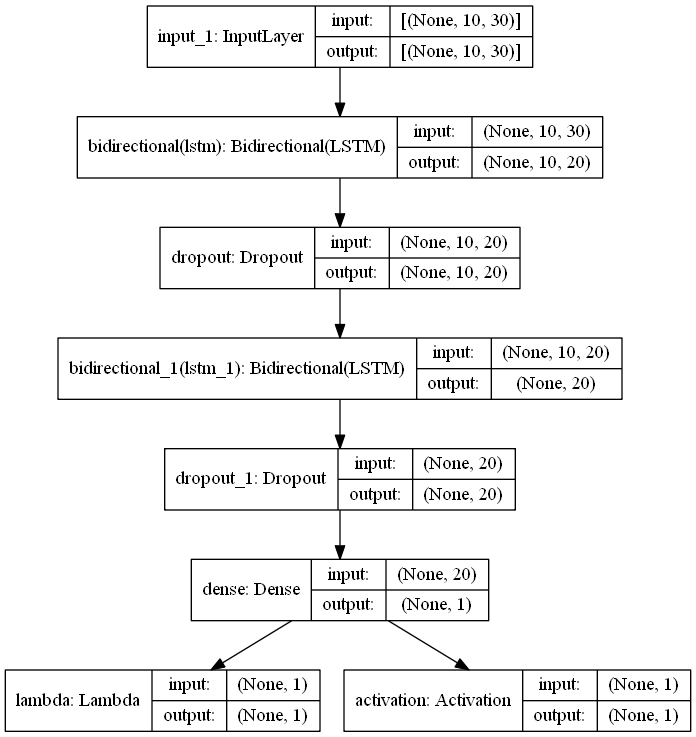}} 
    \subfloat[Generator Model]{\includegraphics[width=0.24\textwidth]{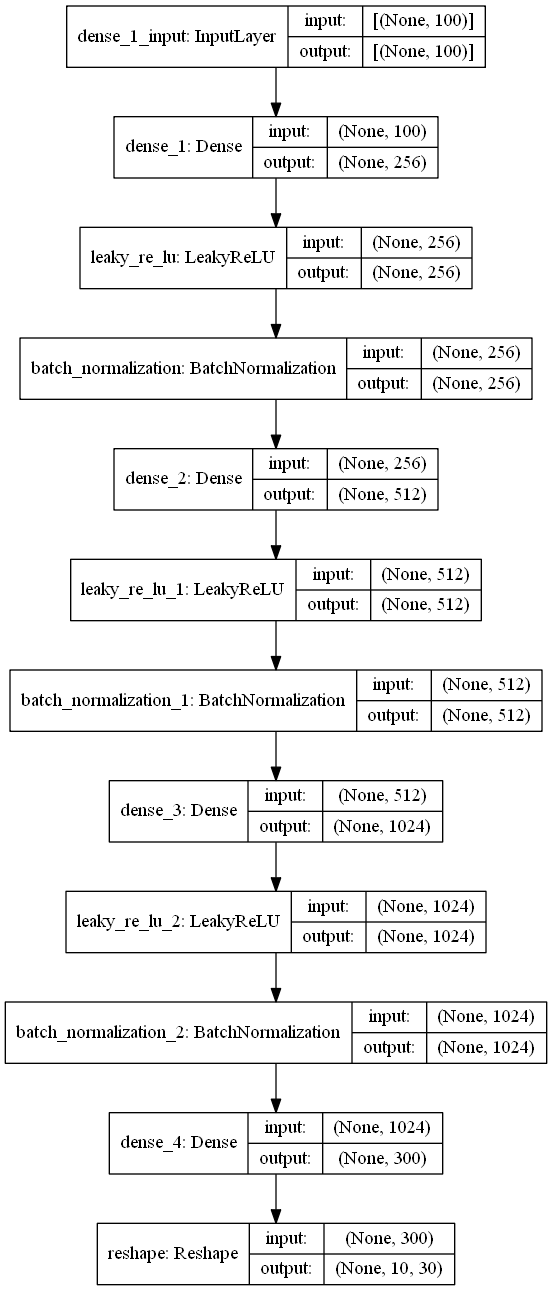}} 
    \subfloat[GAN Model]{\includegraphics[width=0.24\textwidth]{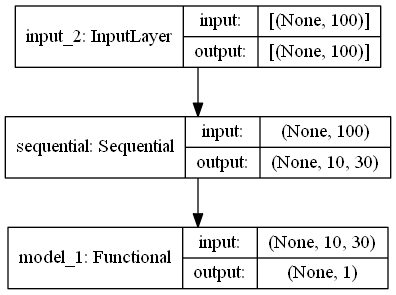}}
    \caption{All of the Neural Networks used in the SGAN model. a is the schema of the combined discriminator and classifier, b is the generator, and c is the connected generative network.}
    \label{fig:sgan_models}
\end{figure}

The SGAN model is an extension of the Generative Adversarial Network for semi-supervised learning. In the traditional GAN architecture, the discriminator is used to determine whether an input is from the generator or another distribution (real or fake images). The model is trained using unlabeled data. Some research involves using transfer learning with the trained discriminator as a starting point when developing a classifier for the same dataset. Therefore the discriminator is can be trained using both supervised and unsupervised techniques. Many areas of research use this technique with promising results \cite{odena2016semi, madani2018semi, salimans2016improved}. 

In the SGAN extension, the discriminator is used to predict the number of classes in the problem along with an extra output to identify whether the data in from the real or generated distribution. The training is done with both supervised and unsupervised:

\begin{itemize}
  \item Unsupervised: Trained the same way at the traditional GAN. It identifies if the input is from a real or fake distribution.
  \item Supervised: Trained by predicting what class the input is from.
\end{itemize}

In the implementation of the SGAN model for this paper, the unsupervised (Discriminator, D) and the supervised (Classifier, C) learning is done in the same network. Both C and D share the same weights except for the output layer \cite{odena2016semi}. The D model is trained by passing in generated data (created using generator \ref{fig:sgan_models} b) or from actual physiological data. The C model is trained purely on the labeled data. The Generator is trained using traditional GAN techniques.

\begin{table}
\centering
\resizebox{\columnwidth}{!}{%
\begin{tabular}{lrrrrrrrrrr}
\toprule
{} &      Loss &   Accuracy &       AUC &      TP &      TN &     FP &          FN &  Precision &    Recall &        F1 \\
\midrule
S2  &  0.516924 &  0.945650 &  0.965004 &  2379.3 &  2097.8 &  217.0 &   40.7 &   0.917598 &  0.983196 &  0.676486 \\
S3  &  0.513025 &  0.946733 &  0.993501 &  2305.7 &  2339.2 &   38.0 &  223.3 &   0.984374 &  0.911691 &  0.680289 \\
S4  &  0.517611 &  0.928677 &  0.972539 &  2410.0 &  2150.4 &  253.2 &   97.0 &   0.908693 &  0.961308 &  0.675961 \\
S5  &  0.512632 &  0.931547 &  0.980149 &  2497.3 &  2127.1 &  285.9 &   53.7 &   0.899272 &  0.978963 &  0.678918 \\
S6  &  0.526812 &  0.931950 &  0.974355 &  2534.9 &  2077.9 &  299.9 &   37.1 &   0.895664 &  0.985571 &  0.683887 \\
S7  &  0.492325 &  0.983055 &  0.998540 &  2460.7 &  2391.0 &   15.6 &   68.3 &   0.994087 &  0.972980 &  0.677625 \\
S8  &  0.518550 &  0.929580 &  0.957873 &  2600.1 &  2114.9 &  298.0 &   58.9 &   0.897414 &  0.977853 &  0.687908 \\
S9  &  0.508347 &  0.918945 &  0.993619 &  2172.6 &  2372.8 &   23.3 &  378.4 &   0.989712 &  0.851649 &  0.680447 \\
S10  & 0.501299 &  0.963571 &  0.993149 &  2867.1 &  2250.8 &  162.1 &   30.9 &   0.948772 &  0.989341 &  0.706083 \\
S11  &  0.510711 &  0.899842 &  0.983076 &  2349.9 &  2228.8 &  156.6 &  353.1 &   0.945005 &  0.869363 &  0.693859 \\
S13 &  0.509713 &  0.884827 &  0.929205 &  2632.7 &  1840.4 &  581.7 &    0.3 &   0.819442 &  0.999873 &  0.684961 \\
S14 &  0.509615 &  0.930123 &  0.984960 &  2368.3 &  2366.0 &   43.1 &  312.7 &   0.983424 &  0.883377 &  0.689998 \\
S15 &  0.567038 &  0.890280 &  0.940721 &  2574.1 &  2007.3 &  409.6 &  154.9 &   0.865780 &  0.943243 &  0.693094 \\
S16 &  0.579891 &  0.931138 &  0.961493 &  2667.2 &  2055.9 &  344.4 &    4.8 &   0.888003 &  0.998212 &  0.690064 \\
S17 &  0.592262 &  0.830760 &  0.891321 &  2394.6 &  1890.6 &  378.6 &  494.4 &   0.866081 &  0.828853 &  0.718030 \\
\bottomrule
\end{tabular}
}
\caption{Results of the fully supervised learning algorithm,}
\label{table:super}
\end{table}

\begin{table}
\centering
\resizebox{\columnwidth}{!}{%
\begin{tabular}{llllllllllr}
\toprule
{} &      Loss &  Accuracy &       AUC &      TP &      TN &      FP &     FN & Precision &    Recall &        F1 \\
\midrule
S2  &  0.689051 &  0.591747 &  0.559218 &  1812.0 &  1013.0 &  1341.0 &  608.0 &  0.574691 &   0.74876 &  0.672783 \\
S3  &  0.645258 &  0.666326 &  0.817061 &  2445.0 &   814.0 &  1548.0 &   84.0 &  0.612322 &  0.966785 &  0.681671 \\
S4  &   0.63233 &  0.667944 &  0.891278 &  2507.0 &   804.0 &  1646.0 &    0.0 &   0.60366 &       1.0 &  0.671758 \\
S5  &  0.694463 &  0.615307 &  0.488646 &  2307.0 &   756.0 &  1671.0 &  244.0 &   0.57994 &  0.904351 &  0.677646 \\
S6  &  0.652812 &  0.666065 &  0.762319 &  2512.0 &   811.0 &  1606.0 &   60.0 &  0.610005 &  0.976672 &  0.680333 \\
S7  &  0.671676 &  0.762292 &  0.687121 &  2213.0 &  1539.0 &   854.0 &  316.0 &  0.721552 &  0.875049 &  0.678835 \\
S8  &  0.626768 &  0.810656 &  0.896574 &  1756.0 &  2337.0 &    53.0 &  903.0 &  0.970702 &  0.660399 &  0.689933 \\
S9  &  0.532142 &  0.903128 &  0.928654 &  2463.0 &  1984.0 &   389.0 &   88.0 &  0.863604 &  0.965504 &  0.682542 \\
S10 &  0.634894 &  0.828268 &  0.830461 &  2685.0 &  1757.0 &   708.0 &  213.0 &  0.791335 &  0.926501 &  0.701610 \\
S11 &   0.68506 &    0.6593 &   0.57124 &  2703.0 &   668.0 &  1742.0 &    0.0 &  0.608099 &       1.0 &  0.691658 \\
S13 &  0.663729 &  0.727543 &  0.753587 &  2633.0 &  1036.0 &  1374.0 &    0.0 &    0.6571 &       1.0 &  0.686034 \\
S14 &  0.596118 &  0.879513 &  0.947014 &  2569.0 &  1913.0 &   502.0 &  112.0 &  0.836535 &  0.958225 &  0.689469 \\
S15 &  0.566948 &  0.814684 &  0.860171 &  2729.0 &  1421.0 &   944.0 &    0.0 &  0.742989 &       1.0 &  0.697686 \\
S16 &  0.665796 &  0.667849 &  0.651586 &  2050.0 &  1338.0 &  1063.0 &  622.0 &  0.658529 &  0.767216 &  0.689994 \\
S17 &  0.608933 &  0.735397 &  0.831531 &  2856.0 &   921.0 &  1326.0 &   33.0 &  0.682927 &  0.988577 &  0.720000 \\
\bottomrule
\end{tabular}
}
\caption{Results of the semi-supervised learning algorithm,}
\label{table:sgan}
\end{table}

\begin{figure}[ht]
\includegraphics[scale=0.4]{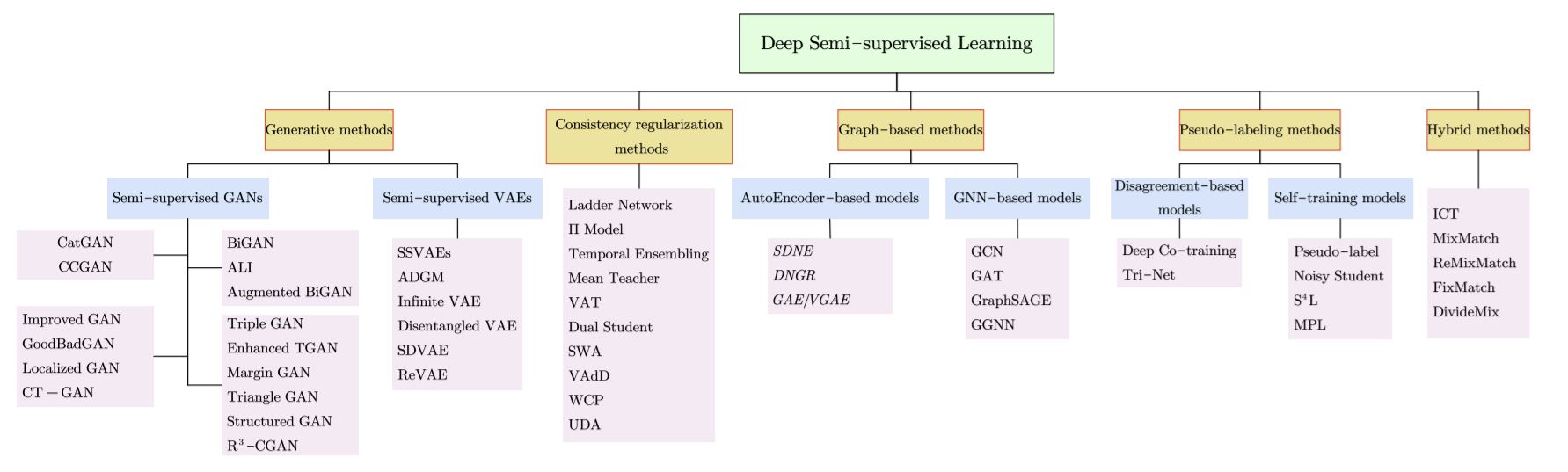}
\centering
\caption{Survey of common semi-supervised learning techniques \cite{Yang2021}}
\label{fig:all-semi}
\end{figure}

\subsection{Model Training and Evaluation Details}
When testing and training the models, it was clear that the ratio of positive to negative labels was heavily skewed. 75\% of the data had a negative label while the rest had a positive label. Training on this split led to models failing too often (model predicted only one class). So two techniques were considered;

\begin{itemize}
  \item Duplicating the positive labels
  \item Dropping negative samples using a normal distribution
\end{itemize}

Both the techniques worked well, but the second option was chosen just to not allow repeated training samples. In practice, the second option works faster than the first technique since there is less data to train with.

All the model in this paper were evaluated using the following metrics: Loss, Binary Accuracy, AUC, True Positive, True Negative, False Positive, False Negative, Precision, Recall, and F1 Score. When explicitly comparing how well a model performed over another, Binary Accuracy and AUC were the two most important metrics.

When testing the models, a Leave-One-Subject-Out (LOSO) method was used. 1 subject out of the 15 was taken as the test subject and the model was trained using the other 14 subject. The evaluation and comparison was done using the testing sample. 

It should also be noted there is no hyper-parameter tuning in this paper.

\section{Results}

Table \ref{table:super} shows the results of the fully supervised algorithm using LOSO with the metrics described above. The algorithm was testing 50 times and the table shows the mean of the metrics for all tests. While the accuracy of the subjects have a range, it is clear that the model is able to solve the problem. Both the binary accuracy and the AUC metric confirm the success of the fully supervised algorithm. 

Table \ref{table:sgan} shows the results of the semi-supervised learning algorithm. The algorithm was also tested 50 times to generate the results. The same metrics show that the model was able to learn. But it is clear that the model was no where near as successful as the fully supervised learning technique. Only 30\% of the data was used a labeled and the rest was unlabeled. However, it shows that the SGAN model is able to learn and is relatively close in performance with certain subjects. 

\section{Discussions}

The results of the SGAN model show that there is potential with SSL and physiological data. This work is in line with other researchers who used SSL to work with EEG data for emotion classification. However, there was nothing in literature about stress classification using Empatica E4 data for stress classification using Semi-Supervised Learning. 

This work accomplished its task of showing the potential of SSL and comparing Bi-LSTM and SGAN. The findings of this paper show expensive labeling of physiological can be made cheaper by using SSL. SSL allows for research to focus on improving Human-Computer interactions without having to spend considerable amount of time of with labeling. Many fields that use ML with physiological data can benefit from SSL as it lowers the expenses of labeling. 

Initially, the paper aimed to see if SSL algorithms work with the E4 sensor for stress classification. While the results of this paper is promising, more research in different avenues must be conducted. Future works to confirm the effectiveness of SSL includes:

\begin{itemize}
  \item The SGAN algorithm used in this paper aims to solve the problem using 30\% labeled data and 70\% unlabeled data. It is worth considered that the models might have learned only by using the labeled data and ignoring the unlabeled data. So a fully supervised algorithm trained on the same amount of labeled data must be created to test the effectiveness of the SGAN model. 
  \item Different ratios of labeled and unlabeled data must be tested to see how the accuracy of the model changes depending on the amount of labeled data. This would also provide us with a way of gauging what number of labeled data to use to lower cost and maximize accuracy. 
  \item Different SSL techniques must be tested to see which ones perform well on physiological data. Figure \ref{fig:all-semi} shows a collection of potential semi-supervised techniques that can be tested. 
  \item Hyper parameters need to be tuned using techniques such as Bayesian Optimization \cite{snoek2012practical}.
\end{itemize}

\section{Conclusion}
SSL is a method that allows for training a model on both a limited amount of expensive labeled data and abundantly available unlabeled data. This paper focuses on comparing a supervised algorithm and a SSL algorithm on physiological data. Current research focuses on using a fully labeled dataset for supervise learning which is expensive labeling process. SSL algorithms would allow for cheaper emotion or stress classification models using physiological data. This paper provides a comparison of a fully supervised and semi-supervised algorithms for stress classification, and shows its effectiveness. Also provided are design elements such as using CNN or LSTMs and pre-processing data. Finally, the limitations of this paper are highlighted and potential research avenues to expand and confirm the findings of this paper are presented. 

\bibliographystyle{unsrt}  
\bibliography{references}  


\end{document}